# Integrating Activity Predictions in Knowledge Graphs

Alec Sculley[1, *], Cameron Stockton[1] and Forrest Hare[1]

[1] *Summit Knowledge Solutions, Arlington, Virginia, United States of America*

**Abstract**

We argue that ontology-structured knowledge graphs can play a crucial role in generating predictions about future events. By leveraging the semantic framework provided by Basic Formal Ontology (BFO) and Common Core Ontologies (CCO), we demonstrate how data—such as the movements of a fishing vessel—can be organized in and retrieved from a knowledge graph. These query results are then used to create Markov chain models, allowing us to predict future states based on the vessel's history. To fully support this process, we introduce the term `spatiotemporal instant' to complete the necessary structural semantics. Additionally, we critique the prevailing ontological model of probability, according to which probabilities are about the future. We propose an alternative view, where at least some probabilities are treated as being about actual process profiles, which better captures the dynamics of real-world phenomena. Finally, we demonstrate how our Markov chain-based probability calculations can be seamlessly integrated back into the knowledge graph, enabling further analysis and decision-making.

## 1. Introduction

Often people wonder what the probability is that some event might occur in the near future. We ask, for example, what the chance is that it will rain tomorrow, and we ask about the likelihood of an imminent economic recession. In order to answer these questions, we take into account the conditions that precede rainy days and recessions. The goal of the calculations is to plan an efficient use of resources: if there is a 90% chance of rain tomorrow, we bring an umbrella; if there is a high probability of a recession, we shore up our investments.

Prior to the use of knowledge graphs, e.g. relational databases structured by general data models, these probabilities would be the result of a calculation whose input data was collected, or structured, for the sole purpose of determining the chance of rain or a recession. Knowledge graphs facilitate the collection and organization of information, and the querying of that information for any analysis of the data that can benefit from the logical structure of the knowledge graph. This database structure offers a clear advantage: information needs only to be collected and structured once, but can be used to answer any number of questions without significant restructuring. Such uses include answering questions about the probabilities of possible events.

Ontologies - logically structured vocabularies - give structure and meaning to the information in the knowledge graphs. Ontologies therefore allow for the integration of data collected from disparate databases and different schemas. Imagine these organizational and computational issues when stored information about rain and economic conditions is contained in disparate databases organized according to different schemas. A knowledge graph structured by an ontology allows for integration and analyses of the probabilities of these events regardless of the structure of the source databases.

* Corresponding author.

alec@sks.ai (A. Sculley); cameron@sks.ai (C. Stockton); forrest@sks.ai (F. Hare)

0000-0002-0369-1965 (A. Sculley); 0000-0001-5655-9119 (F. Hare)

This paper provides a general way to use knowledge graphs and Markov analyses to support queries that return the probabilities of future events. Section II explores the methods used to acquire and store the data for input into the Markov analyses. Section III demonstrates a first-order Markov analysis. Section IV demonstrates a second-order Markov analysis. Section V discusses how the results of the Markov analyses can be integrated back into the knowledge graph for additional uses. Section VI presents future directions of research. Section VII offers conclusions.

## 2. Method

### 2.1. Basic Formal Ontology and the Common Core Ontologies

Basic Formal Ontology (BFO) is a top-level ontology, which means it is a domain neutral representation of reality at its most general [1]. Everything in BFO is an entity or a relationship that exists between entities. Entities are things like baseballs, plans, oceans, and the process of a fishing trip. BFO accounts for entities in terms of continuants and occurrents, which are distinguished by their relationships to time. Continuants continue through time, which means that continuants lack temporal parts (they lack a duration, for example) and exist through time. Examples are a particular baseball, the President of the United States, and the sun. Continuants contrast with occurrents, like processes, that have a beginning and ending. A particular fishing trip is an example of an occurrent. It has a start and end time - it occurs in time, but does not continue through time.

Both BFO and CCO are realist ontologies, which means that they are models of reality according to subject matter experts [3]. In many cases, subject matter experts are scientists, but in other cases, they are the data stakeholders who have a privileged understanding of the domain they are interested in.

In this paper, we shall use BFO and CCO to model the following hypothetical case:

> *There is a fishing vessel of interest to fisheries ecosystem management analysts that is at some location. After observing the fishing vessel for 100 days and collecting information about the fishing vessel's movements, analysts correctly conclude that each day the fishing vessel either stays at its location or travels to one of two other fixed locations.*

Ultimately we want to answer the following question:

> *Based only on previous behavior of the fishing vessel, what is the probability that the fishing vessel will travel to locationX the day after it travels to locationY?*

'LocationX' and 'locationY' in this question are each intended to be interchangeable with 'location1', 'location2', and 'location3'. As such, 'travel to' is intended to be read as accounting for cases where the vessel goes to a different location, and cases where the vessel remains at the same location.

In order to model this case, and ultimately write a SPARQL query for our question, we focus on the classes in Table I, and properties in Table II, all of which are in BFO [1] and CCO [2].

**Table I**

Definitions or elucidations of classes

| Identifier | Definition or Elucidation |
|---|---|
| bfo:Spatiotemporal Region | A spatiotemporal region is an occurrent that is an occurrent part of spacetime. |
| bfo:Spatial Region | A spatial region is a continuant entity that is a continuant part of the spatial projection of a portion of spacetime at a given time. |
| bfo: Temporal Region | A temporal region is an occurrent over which processes can unfold. |
| bfo:Temporal Instant | A temporal instant is a zero-dimensional temporal region that has no proper temporal part. |
| bfo:Process | p is a process means p is an occurrent that has some temporal proper part and for some time t, p has some material entity as participant. |
| bfo:Process Boundary | A temporal part of a process that has no proper temporal parts. |
| bfo:History | A history is a process that is the sum of the totality of processes taking place in the spatiotemporal region occupied by the material part of a material entity. |
| bfo:Process Profile (curated in CCO) | An occurrent that is an occurrent part of some process by virtue of the rate, or pattern, or amplitude of change in an attribute of one or more participants of said process. |
| cco:Watercraft | A Vehicle that is designed to convey passengers, cargo, or equipment from one location to another by water travel. |
| cco:Probability Measurement Information Content Entity | A Measurement Information Content Entity that is a measurement of the likelihood that a Process or Process Aggregate occurs. |
| cco:Vehicle Track Point | An Object Track Point that is where a Vehicle is or was located during some motion. |
| bfo:Disposition | A disposition b is a realizable entity such that if b ceases to exist then its bearer is physically changed & b's realization occurs when and because this bearer is in some special physical circumstances & this realization occurs in virtue of the bearer's physical make-up. |

**Table II**
Definitions or elucidations of properties

| **Identifier** | **Definition or Elucidation** |
|---|---|
| bfo:Precedes | Precedes is a relation between occurrents o, o' such that if t is the temporal extent of o & t' is the temporal extent of o' then either the last instant of o is before the first instant of o' or the last instant of o is the first instant of o' & neither o nor o' are temporal instants. |
| cco:is a measurement of | x is_a_measurement_of y iff x is an instance of Information Content Entity and y is an instance of Entity, such that x describes some attribute of y relative to some scale or classification scheme. |
| cco:has datetime value | A data property that has as its value a datetime value. |
| cco:measurement annotation | A measurement value of an instance of a quality, realizable or process profile. |
| Bfo:spatially projects onto | Spatially projects onto is a relation between some spatiotemporal region b and spatial region c such that at some time t, c is the spatial extent of b at t. |
| Bfo:temporally projects onto | Temporally projects onto is a relation between a spatiotemporal region s and some temporal region which is the temporal extent of s. |
| Bfo:participates in | Participates in holds between some b that is either a specifically dependent continuant or generically dependent continuant or independent continuant that is not a spatial region & some process p such that b participates in p some way. |
| Bfo:inheres in | b inheres in c =Def b is a specifically dependent continuant & c is an independent continuant that is not a spatial region & b specifically depends on c. |
| Bfo:realizes | Realizes is a relation between a process b and realizable entity c such that c inheres in some d & for all t, if b has participant d then c exists & the type instantiated by b is correlated with the type instantiated by c. |
| Bfo:occupies spatial region | b occupies spatial region r =Def b is an independent continuant that is not a spatial region & r is a spatial region & there is some time t such that every continuant part of b occupies some continuant part of r at t and no continuant part of b occupies any spatial region that is not a continuant part of r at t. |
| Bfo:occupies spatiotemporal region | Occupies spatiotemporal region is a relation between a process or process boundary p and the spatiotemporal region s which is its spatiotemporal extent. |
| Bfo:spatial part of | x spatial part of y iff x, y, z, and q are instances of Immaterial Entity, such that for any z connected with x, z is also connected with y, and q is connected with y but not connected with x. |
| Bfo:occurrent part of | Occurrent part of is a relation between occurrents b and c when b is part of c. |
| Bfo:temporal part of | b temporal part of c =Def b occurrent part of c & (b and c are temporal regions) or (b and c are spatiotemporal regions & b temporally projects onto an occurrent part of the temporal region that c temporally projects onto) or (b and c are processes or process boundaries & b occupies a temporal region that is an occurrent part of the temporal region that c occupies). |
| Bfo:history of | History of is a relation between history b and material entity c such that b is the unique history of c. |

## 2.2. Getting data into the knowledge graph

Since this is a hypothetical case, we created a set of dummy data that includes the randomly generated locations of a fishing vessel of interest over one hundred days. The fishing vessel can be located at location1, location2, or location3, at any day, and, once per day, the vessel either stays where it is or moves to a different location. Table III shows the first three days, and the last two days of data.

**Table III**
Sample data

| Time | Day | Location |
|---|---|---|
| 2023-04-08 12:00:00 | Day1 | location3 |
| 2023-04-09 12:00:00 | Day2 | location1 |
| 2023-04-10 12:00:00 | Day3 | location3 |
| 2023-07-15 12:00:00 | Day99 | location1 |
| 2023-07-16 12:00:00 | Day100 | location1 |

To create our knowledge model, we used an ontology development tool with a plug-in that ingests spreadsheet data. Additional data was added to the spreadsheet in order to more efficiently use the plug-in to ingest our data into a BFO conformant knowledge graph. This includes columns for instances of fishing trip, spatiotemporal region, spatial region, and temporal region, as well as a column for a single instance of fishing vessel. In the interest of space, we do not show the full spreadsheet here.

We refer to the fishing vessel of interest as “fishingVessel”. FishingVessel is ingested as an instance of Watercraft and ingested as participating in a single fishing trip, which is a Process that we refer to as “fishingTrip”.

FishingTrip is identified as occupying some Spatiotemporal Region, which temporally projects only onto Days 1-100. FishingTrip is codified as having one hundred occurrent parts that signify the activities of each day of the trip. All one hundred parts of fishingTrip are ingested as instances of Process. Each part of fishingTrip bfo:precedes and bfo:is_preceeded_by some other part of fishingTrip, except for the temporally first and last parts of fishingTrip, which, respectively, only bfo:precede or only bfo:is_preceeded_by some part of fishingTrip. Each part of fishingTrip is ingested with a single occurrent part which is the occurrent during which fishingVessel undergoes observation. The occurrent parts of the occurrent parts of fishingTrip are ingested as instances of Process Boundary.

Process boundaries lack proper temporal parts, so each process boundary instance occupies an instance of Spatiotemporal Instant. A *spatiotemporal instant* is a spatiotemporal region that spatially projects onto a zero-dimensional spatial region and temporally projects onto a temporal instant at the same moment in time. In other words, it is a spatiotemporal region without some spatiotemporal region as a proper part. We the term ‘spatiotemporal instant’ in order to complete the structural semantics necessary for modelling our use case.

We are ultimately interested in fishingVessel’s location at specific points in time, so we created one-hundred instances of Temporal Instant. Each temporal instant is the temporal projection of a spatiotemporal instant. We ingested each time in the spreadsheet as a datetime value of some temporal instant.

We also created a single instance of Vehicle Track Point, which we refer to as “vesselTrackPoint”. VesselTrackPoint is the zero-dimensional spatial region that fishingVessel01 always occupies at some discrete time. Ultimately, we can find out fishingVessel’s location at a time by determining what spatial region has vesselTrackPoint as a part at that time.

Instances of location are ingested into the graph as instances of spatial region.[2]

Now that we have specified the structure of our application ontology and modeling conventions, Fig. 1 is the resulting graph for randomly chosen day_62.

[2] Since locations can be absolutely measured in relation to the center of the Earth’s geoid, it is appropriate to use bfo:spatial_region here.

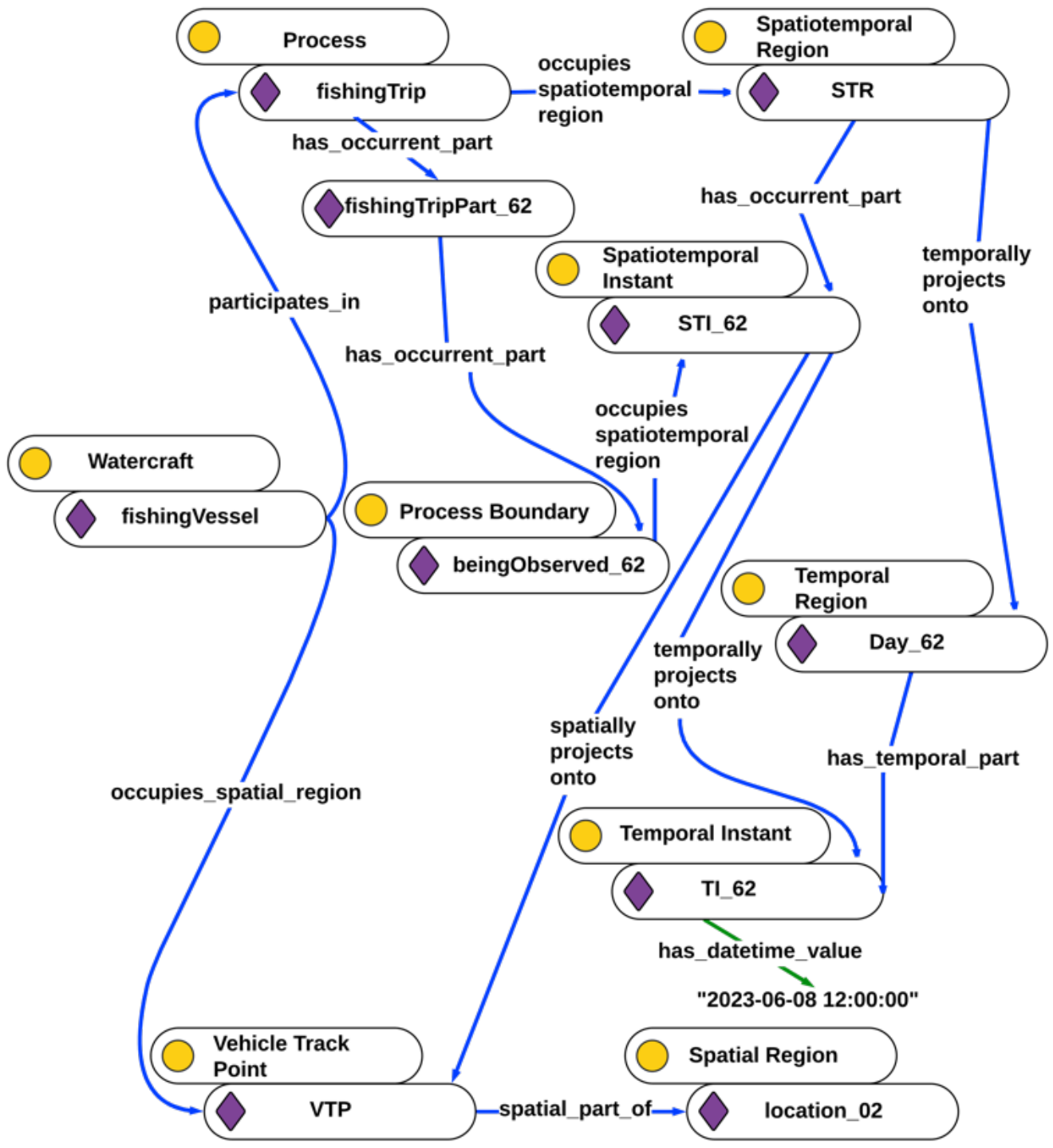

**Figure 1:** Day 62 Graph of fishingVessel

We want to use the structure of the graph in Fig. I to allow us to calculate the probability that the ship of interest will travel to locationX after locationY. It's important to note that Fig. 1 is only a snapshot of reality - at one particular temporal instant. A full graph would connect fishing trips at different temporal instants using the bfo:precedes object property, as noted earlier.

## 2.3. Queries and results

One way to return the details of each state transition of fishingVessel is to write a query that leverages the datetime values to return locations in order.

```
SELECT ?datetime ?location
WHERE
{
fishingVessel bfo:occupies_spatial_region ?fishingVesselTrackPoint .
?fishingVesselTrackPoint cco:spatial_part_of ?location .
?spatiotemporalInstant Bfo:spatially_projects_onto ?fishingVesselTrackPoint
.
?spatiotemporalInstant Bfo:temporally_projects_onto ?temporalInstant .
?temporalInstant cco:has_datetime_value ?datetime .
}
ORDER BY ?datetime
```

This query gives us results that may look like Table IV. In this case, the query looks at the locations of fishingVessel at every time period in question.

**Table IV**

FishingVessel locations from 8-10 April 2023

| datetime | location |
|---|---|
| 2023-04-08 12:00:00 | location3 |
| 2023-04-09 12:00:00 | location1 |
| 2023-04-10 12:00:00 | location3 |

In this case, the query looks at the locations of fishingVessel at every time in question. Another sort of query returns results across times. It does so by leveraging 'bfo:precedes' to return a list of locations as well as the prior locations that fishingVessel occupied.

```
SELECT ?startLocationOfFishingVessel ?endLocationOfFishingVessel
WHERE
{
?fishingTripPart1 bfo:precedes ?fishingTripPart2 .
?fishingTripPart1 bfo:has_occurrent_part ?beingObserved1 .
?beingObserved1 bfo:occupies_spatiotemporal_region ?spatiotemporalInstant1
.
?spatiotemporalInstant1 bfo:spatially_projects_onto ?fishingVesselTrack-
Point1 .
?fishingVesselTrackPoint1 bfo:spatial_part_of ?startLocationOfFishingVessel
.
?fishingTripPart2 bfo:has_occurrent_part ?beingObserved2 .
?beingObserved2 bfo:occupies_spatiotemporal_region ?spatiotemporalInstant2
.
?spatiotemporalInstant2 bfo:spatially_projects_onto ?fishingVesselTrack-
Point2 . ?fishingVesselTrackPoint2 bfo:spatial_part_of ?endLocationOfFish-
ingVessel .
}
```

This query generates the sort of results in Table V.

**Table V**
Sequence of locations

| startLocationOfFishingVessel | endLocationOfFishingVessel |
|---|---|
| location1 | location3 |
| location1 | location2 |
| location1 | location3 |

After generating the results shown in Table IV and Table V, the outputs of both queries may now be utilized to calculate probability. In the following section, we will demonstrate Markov Chain probabilistic calculations.

# 3. First-Order Markov Chains

## 3.1. Discrete-time Markov chain definition

A Markov chain $X$ is a discrete-time sequence of random variables $X_0, X_1, X_2, \dots$ with values in a finite set $S$, if it follows the Markov property. The Markov property states that, at any time $(t)$, the next state $X_{t+1}$ is conditionally independent of the past $X_0, \dots, X_{t-1}$ given the present state $X_t$ [4]. In a time-homogeneous[3] Markov chain, the transition probabilities do not depend on the time parameter $t$, so the transition matrix remains constant at each step. In this context, each step $t$ represents one day.

The state of the sequence at time $t$ is denoted by a random variable $X_t$, that takes values in $S$. FishingVessel01, has 3 possible locations (states) Thus, our state space may be defined as: $S = \{location1, location2, location3\}$. Note that we are not expanding our state space to capture other possibilities of the vessel's movement. The confined state space will limit the real-world application of these results but allow for a clearer understanding of how to model the resulting probabilities in the knowledge graph.

Moving from one state to another is called a transition. This includes transitions to the same state (often called self-loops). In this way, transition probabilities may be understood as the probabilities of transitioning from one state to another in a single step. We refer to the resultant transition matrix as $P$. It's important to note that transition matrices are an $n\ x\ n$ matrix when the chain has n possible states. The entry $p_{ij}$ represents the probability of transitioning from a state of state-type $i$ to a state of state-type $j$.

Note that in the present context, the relevant state-types (in the parlance of BFO) are occupation-of-location1, occupation-of-location2, and occupation-of-location3. Those types represent the particular states of occupying location1, location2, or location3 that are individuals on particular days. A point about usage: In what follows, for the sake of brevity, we will often use 'location1' to refer not only to the spatial location that is location1 but also to particular states of location1-occupation and to the state-type occupation-of-location1; similarly for 'location2' and 'location3'.

While some may argue that the simplicity of Markov chains limit their applicability, we leverage their simplicity when integrating results into the knowledge graph. The straightforward structure of Markov chains facilitates clear interpretation and updating of data in the model. Markov chains play a foundational role in more complex models, providing us with a robust foundation for more comprehensive analyses where we can incorporate more features and uncertainty.

### 3.2. Determining transition probabilities

We turn our attention to the SPARQL query, which returns a list of 'previousLocation' and 'currentLocation', representing the transitions that FishingVessel01 makes each day. To populate the transition matrix, we sum up each unique transition from state 'previousLocation' to state 'currentLocation.' We divide that number by the total sum of transitions that originated from state 'previousLocation'. For example, there are 9 transitions from location1 to location2. There are 32 transitions originating from location1. The estimated Markov probability[4] of moving to location2, given the present location being location1, as $p_{12} = \frac{9}{32} = 0.281$. This process may be automated using SPARQL queries or Python scripts to efficiently compute transition probabilities.

The resultant matrix P in Table 6 is populated by rows showing present location and columns that reveal next locations for one time step.

**Table VI**

First-order transition matrix

| | **location1** | **location2** | **location3** |
|---|---|---|---|

[3] Time-homogeneity is assumed here for simplicity and practicality. An example of time-inhomogeneity is explained in Section VII(b).

[4] "Markov probability" refers to the probability of moving between states, whilst conforming to the Markov property as stated above.

| | | | |
|---|---|---|---|
| **location1** | 0.375 | 0.281 | 0.344 |
| **location2** | 0.278 | 0.500 | 0.222 |
| **location3** | 0.355 | 0.290 | 0.355 |

### 3.3. How are the first-order Markov probabilities useful?

The transition matrix in Table 6 allows stakeholders to answer questions about future locations of FishingVessel01. For example, given that the vessel of interest is presently at location3 on day 100 (row), we conclude that there is a 29.0% chance that this vessel will be at location2 on day 101 (column).

Now that we have constructed a first order matrix, we may make predictions about the vessel's location beyond only the next day.

**Table VII**
$5^{th}$-step transition matrix

| | **location1** | **location2** | **location3** |
|---|---|---|---|
| **location1** | 0.334 | 0.363 | 0.303 |
| **location2** | 0.334 | 0.363 | 0.303 |
| **location3** | 0.334 | 0.363 | 0.303 |

Recall that the $(i, j)$ entry $p_{ij}^t$ of the transition matrix $P^t$ represents the probability that the Markov chain, starting in a state of state-type $i$, will be in a state of state-type $j$ after $t$ steps. Table 7 shows matrix $P^5$, estimating the probability of the vessel's location after $t = 5$ days.

What if the vessel's movement is more dependent on previously made consecutive steps? Using a higher-order model allows us to capture more complex patterns in the movement.

## 4. Second-Order Markov Chain

### 4.1. First-order v.s. second-order

Second-Order Markov chains function similarly to First-Order Markov chains but with a key difference in how the transitions are determined. In a First-Order Markov chain, the probability of transitioning to the next state depends solely on the present state. However, in a Second-Order Markov chain, the probability of transitioning to the next state depends on both the present state and the previous state. To count as nevertheless adhering to the Markov property, we look at transitions from a state pair $(X_{t-1}, X_t)$. $X_t$ refers to the 'present state', of which the present state is an individual of type $X_t$. $X_{t-1}$ refers to the 'immediate past state', of which the immediate past state was an individual with type $X_{t-1}$. Then, the probability that the entity in question will transition to a state of a given type $X_{t+1}$ is given as follows: $P(X_{t+1}|X_{t-1}, X_t)$.

Including more history when determining future probabilities allows the model to capture more patterns. If a vessel's current movement is influenced by more of its past behavior, a Second-Order Markov chain will capture this with more accuracy.

### 4.2. Determining second-order transition probabilities

To determine the transition probabilities for a Second-Order Markov chain, we look at each possible state pair $(X_{t-1}, X_t)$. Similar to our process in the First-Order Markov chain, we look at the transitions in the historical data from each of the possible state pairs to the following state. For FishingVessel01, there are 3 locations that can be visited. There will be 9 possible state pairs.

We can use SPARQL to retrieve data where each row represents a transition from a specific state pair to a subsequent state. In this format, we can calculate the transition probabilities. What we considered the "present state" in the First-Order Markov chain is now treated as a state pair in the Second-Order model. Because of this, the resulting transition matrix will be larger, reflecting the increased complexity.

### 4.3. How are the second-order Markov probabilities useful?

The Second-Order matrix in Table 8 captures more historical context. This allows us to ask questions such as: Given the vessel was at location1 and is presently in location2, what is the probability that the vessel will move to location3? This can be directly answered from our matrix. In this case, we expect a 22.2% chance of this movement.

**Table VIII**

Second-order transition matrix

| $X_{t-1}$ | $X_t$ | **location1** | **location2** | **location3** |
|---|---|---|---|---|
| **location1** | **location1** | 0.364 | 0.182 | 0.455 |
| **location1** | **location2** | 0.333 | 0.444 | 0.222 |
| **location1** | **location3** | 0.273 | 0.364 | 0.364 |
| **location2** | **location1** | 0.400 | 0.500 | 0.100 |
| **location2** | **location2** | 0.222 | 0.667 | 0.111 |
| **location2** | **location3** | 0.250 | 0.250 | 0.500 |
| **location3** | **location1** | 0.364 | 0.182 | 0.455 |
| **location3** | **location2** | 0.333 | 0.222 | 0.444 |
| **location3** | **location3** | 0.455 | 0.273 | 0.273 |

### 4.4. Possible downsides of a second-order Markov chain

The Second-Order Markov chain provides additional context, allowing for more accurate explanations of processes that rely heavily on patterns.

The primary concern with higher-order models is not the computational complexity, but the requirement of more sufficient historical data. In cases where data is sparse, higher-order Markov models may struggle to capture these sequential dependencies.

The resulting transition matrix will be larger and more complex, increasing the amount of information required to be represented back into the graph. The richer historical context allows us to capture more nuanced relationships and movement patterns, more accurately predicting these state changes in a real-world scenario.

## 5. Updating the Knowledge Graph

This section takes steps toward updating knowledge graphs with probabilities. First, this section presents and assesses the model of probability in CCO. Second, desiderata for a satisfactory model of probability is extracted from the assessment of CCO's model. Third, a new model is developed according to which probabilities are about process profiles.

## 5.1. Probability in the Common Core Ontologies

In CCO, probability is an information content entity. A "Probability Measurement Information Content Entity [(PMICE)]," as probability is labeled, is a "Measurement Information Content Entity that is a measurement of the likelihood that a Process or Process Aggregate occurs" [2]. The process or process aggregate that PMICEs are about can either be past processes or future processes. For example, we can ask what the probability was that a particular asteroid would hit Earth after it safely passes by. Such a measurement is about the past because it is about some process whose time to occur is over. But we can also ask what the probability is that a particular asteroid will hit Earth as it approaches. This is about the future because it is about some process whose time to occur has not begun.

In this paper, interest is in probabilities that inform us about the future, so we focus on the second case where the process' time has not yet begun. The way that the Common Core Ontologies models information that is about future entities is through modal relations. For our purposes, these relations are forward looking, whereas the non-modal versions of the relations are backward looking.

Every Probability Measurement Information Content Entity in CCO is "made in a particular context given certain background assumptions" [2]. This guides us toward defining kinds of Probability Measurement Information Content Entity. For example, a Markov probability, in CCO terms, can be defined as a Probability Measurement Information Content Entity that assumes the Markov Property holds for the entity or entities that it measures. The Markov Property is the property that makes it such that probabilities can be calculated only considering a system's previous state.

Using this model, we can produce the following graph of the probability that fishingVessel at location_01 either goes to location_02, goes to location_03, or stays at location_01.

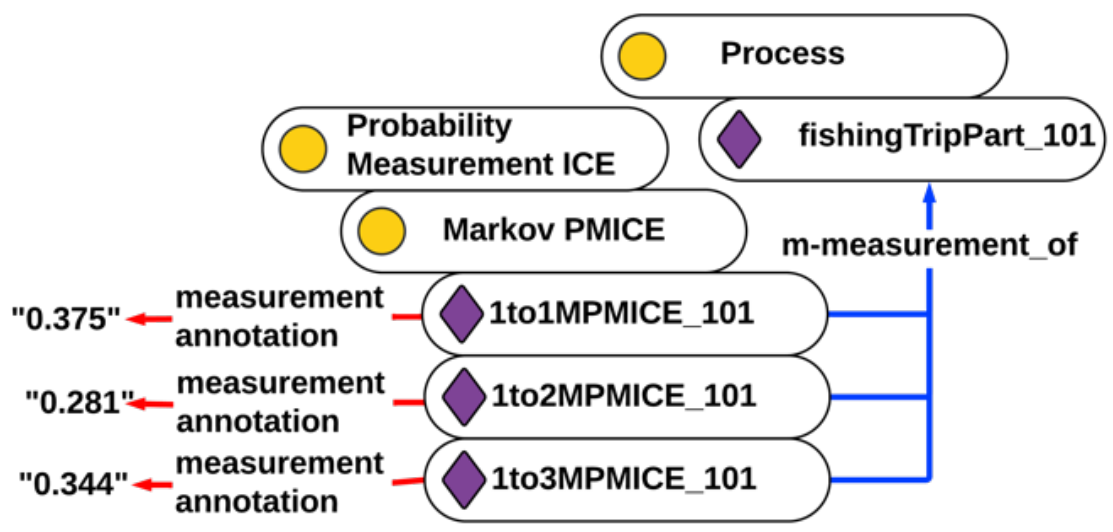

**Figure 2:** Probability of fishingVessel's day 101 location using CCO

In this graph, there are three instances of Markov Probability Measurement Information Content Entity, which correspond to each possible transition in location. Each instance of Markov PMICE is modally about fishingTripPart_101, which is a future process. What the graph tells us is that: there is a 0.375 probability that fishingVessel remains at location01 during fishingTripPart_101; there is a 0.281 probability that fishingVessel travels from location01 to location02 during fishingTripPart_101; and there is a 0.344 probability that fishingVessel travels from location01 to location03 during fishingTripPart_101.

## 5.2. Issues with the CCO model

The CCO model faces the following issues.[5] First, probabilities are not about future processes. Instead, probabilities are about past or present entities that provide us with the ability to make predictions about the future. Second, even if probabilities could be about future processes, the CCO model is silent on what aspects of the future processes the probabilities are about. This becomes a larger issue if we accept that probabilities are not about the future because we then need to know what aspect of past processes probabilities need to be about to allow people to make predictions using them. This subsection spells these issues out in more detail.

[5] In addition to the more substantive issues presented in this subsection, there are legitimate questions about labelling and defining probability-related entities in CCO.

### 5.2.1. Probabilities are not about future processes

This subsection provides motivation to think that probabilities are not about future processes. A full argument for this point will have to wait for a future project. But it is important to provide motivation for the view here because once the connection between probabilities and the future is plausibly severed, we are left with the task of explaining the forward-looking nature of probabilities not about the future. A view that provides such an explanation is a main contribution of this paper.

The view is motivated by arguing (1) statements about the future are not derivable from a set of statements about the past; and (2) probabilities are quotients X/Y such that X and Y are both sets of only statements about the past.

We now propose that a statements about the future – predictive statements – cannot be derived from a set of statements about the past – descriptive statements. The issue here is one of logical validity, which is the property of arguments that makes it such that if all premises are true, the conclusion must be true. The problem is that for any set of descriptive statements with a predictive conclusion, it is possible to deny the conclusion without denying any of the descriptive statements. For example, given the following argument

*(P1) X# of times, S did Y after Z-ing.*
*(P2) A# of times, S did anything after Z-ing.*
*(C1) X/A of the time, S will Y after Z-ing.*

one can deny (C1) and accept (P1) and (P2) without doing anything contradictory. This is because, as far as the argument is concerned, it may be that facts about the past are not the kinds of things that can inform predictions about the future. On the other hand, accepting (P1) and (P2) while denying

*(C2) X/A of the time, S did Y after Z-ing.*

is contradictory. To accept (P1) and (P2) is just to accept (C2).

We shall now show that Markov calculations have the structure of the (P1), (P2), (C2) argument. We will do this through examination of the probability calculations done earlier in this paper. An example Markov probability calculation was described earlier as follows.

> *To populate the transition matrix, we sum up each unique transition from state 'previousLocation' to state 'currentLocation.' For example, there [is a sum of] 9 transitions from location1 to location2.*

This sum is therefore a numerical representation of the nine past state transitions from location1 to location2. But it can be propositionally represented as (P1*) 9 times, fishingVessel travelled from location1 to location2. Next,

> *We divide that number by the total sum of transitions that originated from state 'previousLocation'. [For example,] there are 32 transitions originating from location1.*

This sum is therefore a numerical representation of all the past state transitions that began at location1. But it can be propositionally represented as (P2*) 32 times, fishingVessel travelled from location1 to any location. Next,

*The estimated Markov probability of moving to location2, given the present location being location1, as* $p_{12} = \frac{9}{32} = 0.281.$

The Markov probability is therefore the quotient of two numerical representations of past state transitions. But it can be propositionally represented as (C2*) 9/32 of the time, fishingVessel travelled from location1 to location2. The full propositional representation of the calculation is as follows.

*(P1*) 9 times, fishingVessel travelled from location1 to location2.*
*(P2*) 32 times, fishingVessel travelled from location1 to any location.*
*(C2*) 9/32 of the time, fishingVessel travelled from location1 to location2.*

So, probabilities that are calculated by finding the quotient of two sets of descriptive statements are not about the future. They are about the past. For this paper, we only wish to show that the Markov probabilities of interest have this structure. However, we do think that this argument extends to other kinds of probabilities, including other Markov probabilities, Bayesian probabilities, and so on. This is because, we think, fundamentally all probabilities are quotients of two sets of descriptive statements. If so, no probability is about the future.

However, predictions can be made by naturally assuming that the future will mirror the past as the probability calculations quantify it. The point is that this is a seperate assumption that allows us to make predictions using the probability calculation. It is not, strictly speaking, the probability (i.e., the output of the probability calculation) alone that is a prediction about the future. This is because that assumption does not feature in the probability calculation, or the propositional representation of the probability calculation. As will be shown starting in section 5.2.3, the insights of the current section matter to the ontology modelling of probability.

### 5.2.2. Probabilities are about certain aspects of processes that remain unidentified

Even if probabilities are about future processes, CCO is effectively silent on the aspect of processes that probability measures. CCO says that probability measures the "likelihood that a process occurs," but CCO also says that an alternative label of Probability Measurement Information Content Entity is 'Likelihood Measurement.' Given this, one can only conclude that 'likelihood' and 'probability' are synonymous in CCO.[6] Thus, according to CCO, probability measurements are measurements of probability. This provides no additional information on what probabilities are about.

### 5.2.3. We need to know what probabilities are about

If it is true that probability is not about future processes, then we need to know what it is about. This is for three reasons. One is that because any complete model of probability will have as parts both the information content aspect of probability and an informative model of what that information content measures. Currently, CCO lacks the second part, as shown. The second reason is that we need to know what it is we are basing predictions on. Just saying "probability" is not good enough since it raises the question: "what is the probability based on?" An answer to this question will be explored in the next section. The third, related, reason is that we want to know that we are justified in making predictions based on probabilities. To be justified, the entities that probabilities are about must have some bearing on the future even though they exist in the present.

[6] There is a fine distinction between 'likelihood' and 'probability' in probability theory, but there is not in common sense parlance.

### 5.3. Improving the CCO model

In this section, we improve upon the CCO model of probability. We do so by addressing the issues just mentioned. We show that probabilities are intimately connected to realizable entities that inhere in the participants of processes that we want to make predictions about. This allows us to do the following things: (i) model probability in a way more consistent with what probability calculations are about; (ii) model probability in a way that assists in making predictions about future processes; (iii) understand what probabilities are based on; (iv) understand why predictions based on probability can be justified.

#### 5.3.1. Are probabilities about single realizable entities?

Probabilities are not about future processes, but they may be about parts of past processes that have potential characteristics. A potential characteristic is a characteristic that some continuant has but which can be fulfilled under some set of circumstances. Salt has the potential to dissolve but will not dissolve unless it is placed in the right set of circumstances, like a glass of water. I have the potential to finish writing this paper but will not until I have the correct mindset to fulfill this potential. In BFO, what I have called potential characteristics are called realizable entities. Realizable entities can be realized in processes of a certain type, like being in water, or being focused on finishing writing a paper.

One possibility, then, is that probabilities are about realizable entities. For example, fishingVessel bears three relevant realizables: the realizable to travel from location_01 to location_02; the realizable to travel from location_01 to location_03; and the realizable to stay at location_01. Each instance of Markov PMICE would then be about the correlated realizable entity instance.

This solution is insufficient. Probabilities are not just about single realizable entities. Perhaps they are about realizable entities we want to make predictions about as they compare to the total number of relevant realizable entities that might be realized.

#### 5.3.2. Are all kinds of probabilities about aggregates of realizable entities?

If any probability is about aggregates of realizable entities, it is the probability that a fair six-sided die comes up on one. We intuitively know that there is a 1/6 probability that such a die comes up on one. But why is this the case? The explanation consistent with the view that probabilities are about aggregates of realizable entities is that the die bears six relevant realizables, none of which is in circumstances to increase the chance that it is realized over another relevant realizable. Knowing that the die can only realize one realizable at a time and that there is an aggregate of six realizables that inhere in the die, we get a 1/6 probability. But this explanation does not work for the Markov probabilities that fishingVessel travels to locations 01, 02, or 03. The reason for this is that a Markov probability of a system depends on its previous state and the overall pattern of states of the system over time. Conversely, the probability of a die coming up on one only depends on the aggregate of realizables at a single point in time. So, there are at least some probabilities that are not about aggregates of realizables.

#### 5.3.3. Are some kinds of probabilities about process profiles?

A *process profile* is "an occurrent that is an occurrent part of some process by virtue of the rate, or pattern, or amplitude of change in an attribute of one or more participants of said process" [5]. Some process profiles are magnitudes of changes in attributes of continuants. Call these "magnitude process profiles."[7] Examples are changes of mass, changes in temperature and changes in amounts.

**Table VI**
Magnitude process profiles

| **Process** | **Participant** | **Attribute** | **Magnitude Process Profile** |
|---|---|---|---|
| Losing weight | Person | Weight | Loss of weight |
| Reheating pizza | Pizza | Temperature | Increase in temperature |
| Car racing | Car | Miles travelled | Increase in miles travelled |
| Adding solute to solvent | Solvent | Surface tension | Increase in surface tension of solvent |
| Fishing trip | Boat | Realized Disposition of Location Change | Change in Realized Disposition of Location Change |

There are also process profiles that are abstractions of magnitude process profiles over time. In particular, a *rate process profile* is an occurrent that is an occurrent part of some process by virtue of the rate of change in an attribute of one or more participants of said process. Examples are heartbeat (e.g., beats per minute), speed (e.g., miles per hour) and baseball pitch count average (e.g., pitches per inning).

**Table VII**
Rate process profiles

| **Process** | **Participant** | **Attribute** | **Time Period** | **Rate Process Profile** |
|---|---|---|---|---|
| Heart beating | Heart | Amount of Beats | Minute | Heart rate (beats per minute) |
| Car racing | Car | Miles Travelled | Hour | Rate of speed (miles per hour) |
| Adding solute to solvent | Solvent | Surface tension | Millisecond | Rate of increase in surface tension (millinewton per meter per millisecond) |
| Fishing trip | Boat | Realized Disposition of Location Change | Day | Realized Dispositions of Location Change per Day |

[7] [5] calls these "quality process profiles," but change in some realizable entities, like strength or solubility, can be measured and plotted in a graph in just the same way as change in mass or temperature.

#### 5.3.3.1. Pattern process profiles

An unexplored kind of process profile is the pattern process profile. A *pattern process profile* is an occurrent that is an occurrent part of some process by virtue of the pattern of change in an attribute of one or more participants of said process. For us, patterns are observable regularities in the world. So, a pattern process profile is an occurrent that is an occurrent part of some process by virtue of an observable regularity of change in an attribute of one or more participants in said process.

**Table VIII**
Pattern process profiles

| **Process** | **Participant** | **Attribute** | **Pattern** | **Pattern Process Profile** |
|---|---|---|---|---|
| Orbiting around the Sun | Earth | Climate conditions | Winter, Spring, Summer, Fall | Cyclic |
| Boiling water | Water | Temperature | From room temp, +1-degree F every 5 seconds, until 212 degrees F | Linear |
| Fishing trip | Boat | Realized Disposition of Location Change | 3, 1, 3 . . . 1, 1 | Probabilistic |

#### 5.3.3.2. Pattern of life

A *pattern of life* is an occurrent that is an occurrent part of some process by virtue of the pattern change in realizables that are realized by one or more participants of said process. Thus, patterns of life are pattern process profiles. Examples are an individual's pattern of online activity, an individual's morning routine, and an individual's exercise regimen. Patterns of life need not be restricted to parts of processes that a single individual is an agent in. Indeed, some important patterns of life are parts of processes that groups of people are agents in. The travel pattern of a particular convoy, and the pattern of a guard patrol, are examples.

#### 5.3.3.3. From pattern of life to probability

Patterns of life are often used to determine the probability that some agent will take a future action. If some individual takes route x to work at around 8:30 am and then takes route x (in reverse) home at 5pm, every workday for a year, then there is a very good chance that they will do the same thing on the next workday. There are a couple reasons why an analysis of pattern of life as a pattern process profile allows us to do this. **First**, patterns of life profile realizable entities. In other words, the attributes that patterns of life exist in virtue of are realizable entities – in particular, realizable entities that have been realized in the past. In contrast with the view that probabilities are about aggregates of realizable entities, the process profile view allows us to consider patterns of realized realizables over time. **Second**, the view that patterns of life profile realized realizables over time allows us to explain what probabilities are about, and why we are justified in using them to make predictions. The explanation is that probabilities are measurements of the potentials of realizables. That is, they are measurements of the chance that some realizable will be triggered. However, these measurements are taken by measuring proxies, since potentials are not directly measurable. These proxies are patterns of life.

### 5.3.3.4. Demonstrating the view with the case of fishingVessel

In our case, the pattern of change we are interested in is the change in the pattern of realizables that are realized in transitions between states. In particular, we are interested in the pattern of change in realizables realized in changes in location of fishingVessel. This pattern has already been discussed and used to calculate probabilities earlier in the paper. So, we can straightforwardly use this new construct to model the fishingVessel case.

Recall that the fishingVessel is currently, on day 100, at location 01 and next, on day 101, may either stay at location 01, move to location 02, or move to location 03. Each option requires the fishingVessel to realize a disposition: either the disposition of being at location 01 and remaining at location 01; the disposition of being at location 01 and moving to location 02; or the disposition of being at location 01 and moving to location 03.

Since the fishingTrip is a process that has thus far occurred over 100 days, with the fishingVessel either staying in place, or moving to one of two other fixed locations, a probabilistic pattern of change in realized realizables has been established. This is fishingVessel's pattern of life during fishingTrip (herein we just call this fishingVessel's pattern of life). In the graph this is called "fishingVessel_PoL."

The pattern of life of fishingVessel – fishingVessel_PoL – has parts that we care about more than others. Since the vessel is at location 01, these are the parts relevant to the pattern where the last realization moved fishingVessel to location 01. We can pick these out in the graph by specifying that there occurrent parts of fishingVessel_PoL, namely, 1to1_PoL_Part, 1to2_PoL_Part, and 1to3_PoL_Part. Each part, respectively, is the part of the pattern of change that profiles the realizations of the disposition of staying at location 01, the disposition of moving from location 01 to location 02, and the disposition of moving from location 01 to location 03. See Fig. 3 for a visual representation.

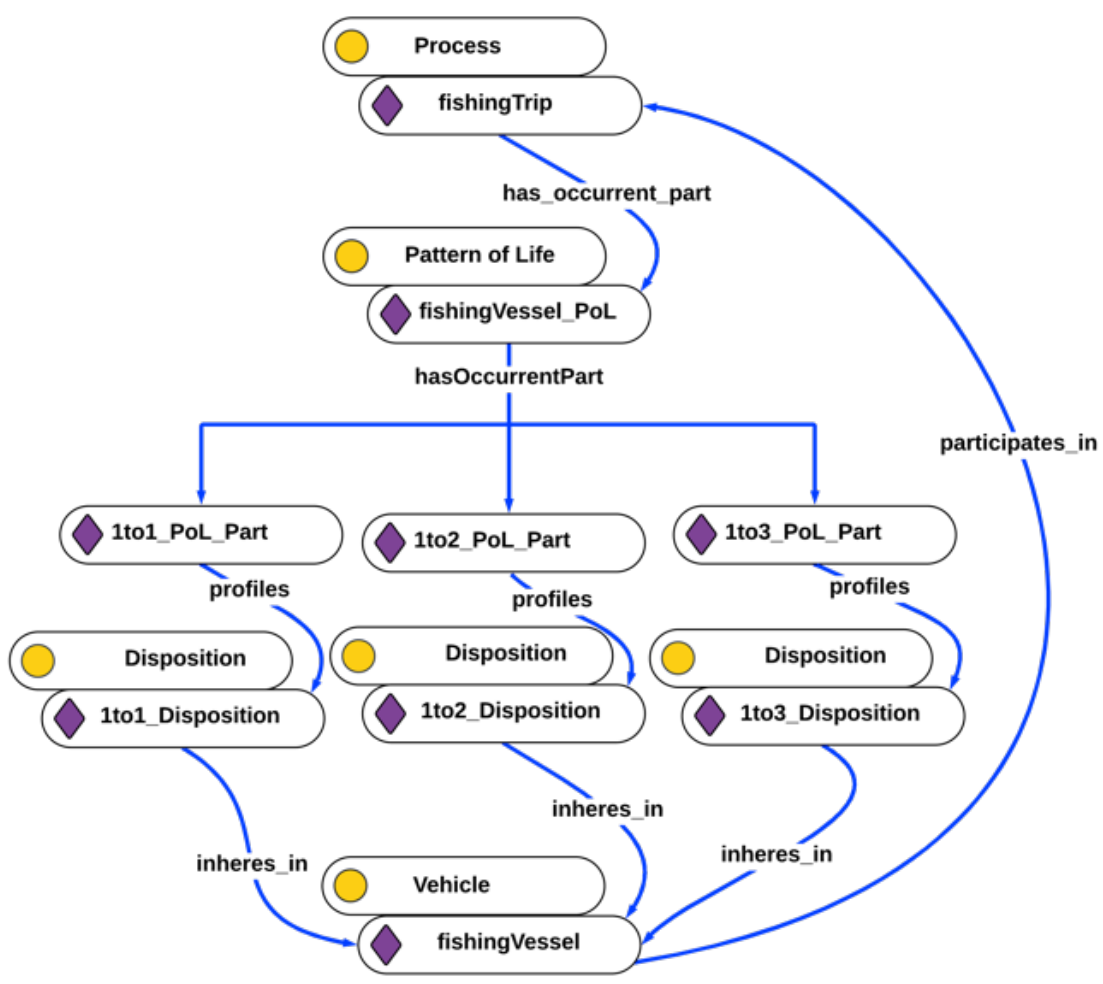

**Figure 3:** FishingVessel's pattern of life and its parts

Next, we want to move from these parts of fishingVessel's pattern of life to the probability values themselves. The first step toward doing this is to count the indivdual times that each disposition is realized in each part of fishingVessel's pattern of life. After, those counts need to be summed to get the total number of times each disposition in question was realized. This is shown visually in Fig. 4. In

Fig. 4: 1to1TransitionCount is the count of times the 1to1_Disposition was realized; 1to2TransitionCount is the count of times the 1to2_Disposition was realized; the 1to3TransitionCount is the count of times that the 1to3_Disposition was realized; and total1toXTransitions is the sum of the three other counts.

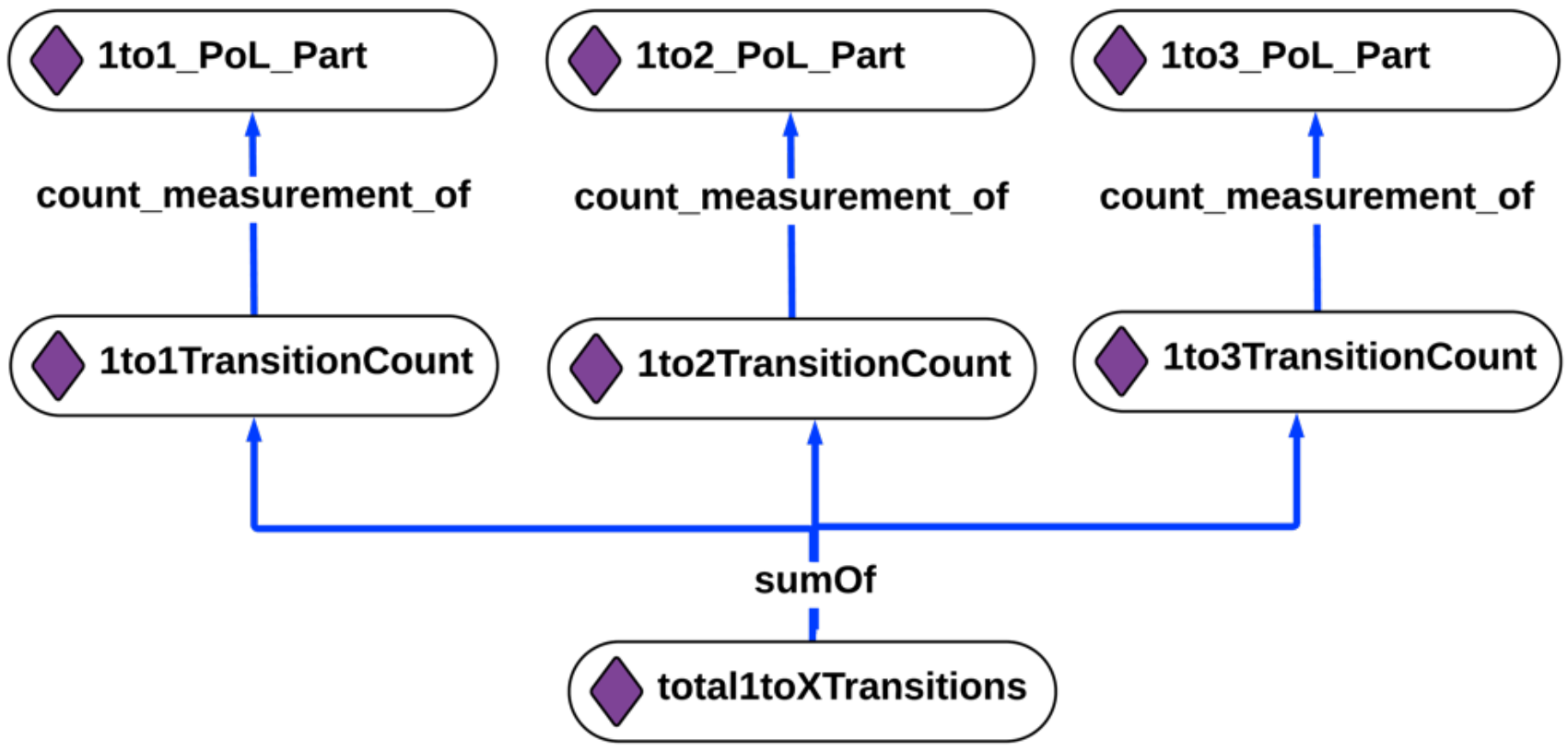

**Figure 4:** Counts of realizations and sum of the counts

Finally, the probability values are reached by putting the count of interest into the numerator of a fraction, and the total count into the denominator of a fraction. These fractions are the Markov Probability Information Content Entities. They are ultimately about fishingVessel's pattern of life, since they are the result of dividing a count that is about the pattern of life.

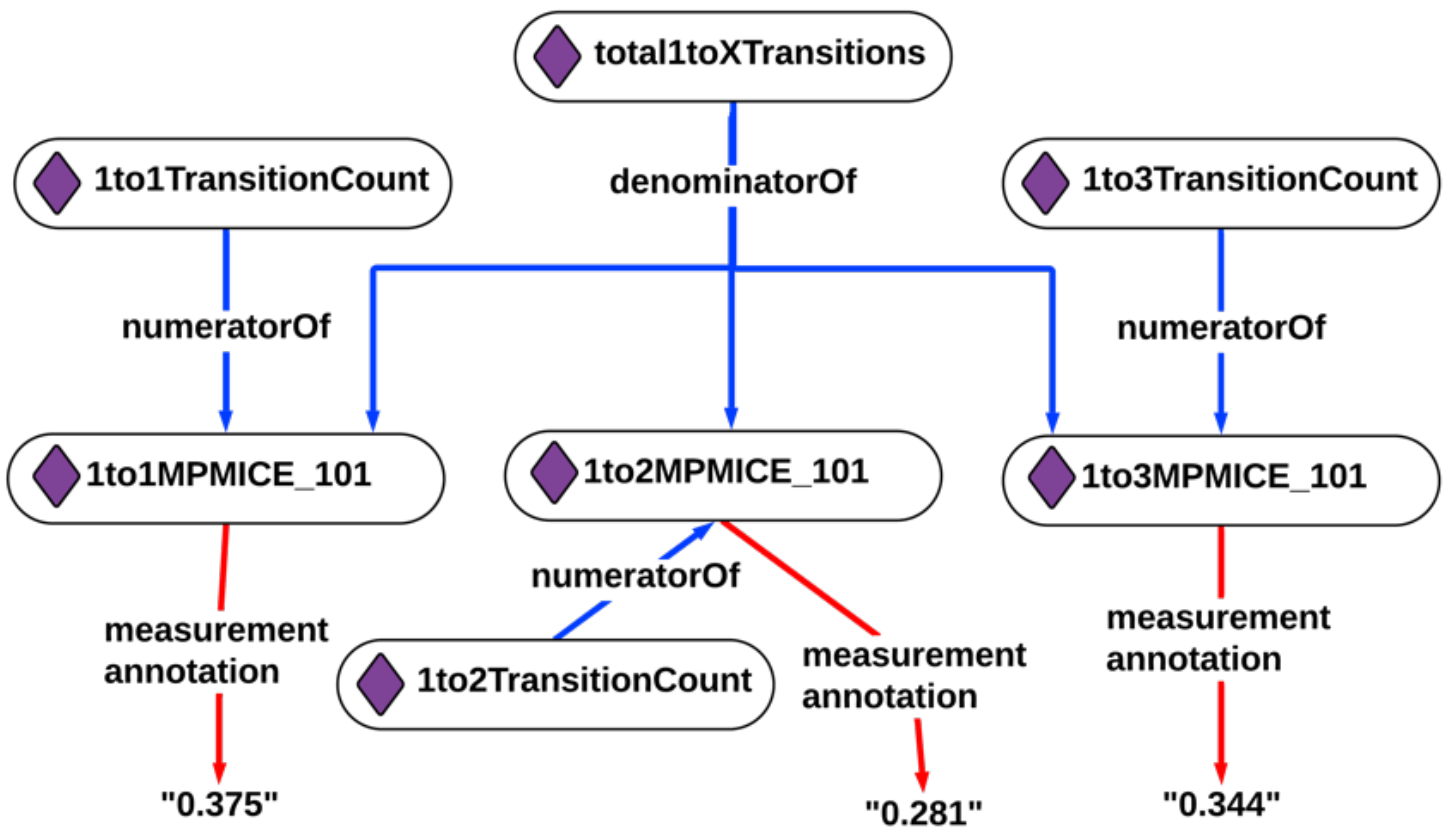

**Figure 5:** Markov probabilities determined by division

## 6. Future Work

### 6.1. Continuous-time Markov chain

In this paper, we've used Discrete-Time Markov Chains (DTMCs), where transitions between states occur at fixed, discrete-time intervals. Recall that the fishing vessel's location was recorded once per day, over 100 days. This approach has advantages, particularly in the stability of transition probabilities over time. Once the transition probabilities are calculated, they do not change over time. Thus, they can be easily stored in the graph, allowing for straightforward predictions regarding the vessel's future movements.

However, this discrete-time model does not capture the randomness of real-world movements. In reality, the movement of a fishing vessel is unlikely to occur at fixed time intervals. Various external factors (weather conditions, fishing regulations, or equipment functionality) could force the vessel to move at any continuous point in time. To model this more realistic behavior, we propose exploring Continuous- Time Markov Chains (CTMCs).

CTMCs allow for transitions between states to occur at any continuous point in time. The time between state changes is modeled using an exponentially distributed random variable and represented in a rate matrix Q. The memoryless nature of the exponential distribution allows CTMCs to adhere to the Markov property. By incorporating time spent at each location in the model, CTMCs provide a more accurate representation of the vessel's movements. For example, if the fishing vessel spends 5 hours in one location before the weather forces it to relocate, CTMCs can capture this.

Despite the advantages, CTMCs introduce some challenges. The transition probabilities are no longer fixed over time. The transition-rate matrix Q describes the instantaneous rate at which the chain transitions between states. From this rate matrix $Q$, we can generate a collection of transition matrices $P(t) = e^{tQ}$. Given that the transition-matrix now depends on $t$ for any future movement of interest, a new matrix will need to be calculated. To store these future probabilities in the graph, the time must be known at which the future event occurred, and the corresponding probabilities are calculated. For any time $t$, a new matrix is required. This greatly increases the complexity of representing such probabilities within the graph, for we now need to predict when we must store probabilities for specific time intervals.

### 6.2. Time-inhomogeneous Markov chain

In this work, the transitions between states were unrealistically calculated without accounting for external factors. One way to account for a more true-to-life case, while using a DTMC, is to allow the transition matrices to vary depending on certain factors. For example, a vessel's movement could vastly differ on weekends vs. weekdays. To address this, we could create unique transition matrices: one representing transitions on weekdays and one representing transitions on weekends. Depending on the day of the week, we can then apply the appropriate matrix to predict the future movement.

This same principle could be applied to other factors, such as fishing regulations, weather seasons, or operational hours.

Incorporating time-inhomogeneity would make the model more dynamic and aligned with real-world variations in the movements.

### 6.3. Discrete locations

An assumption in this example relies on the locations or states to be somewhat general. In real-world scenarios, when observing a vessel, locations may not be recorded in a discrete manner. We therefore would likely see some geo-coordinates that represent the vessel's location at some time. This would fundamentally increase our state space to be somewhat immeasurable and the resultant Markov model would produce somewhat meaningless results.

To deal with geo coordinates, while still implementing a Markov chain, we cluster observations as a pre-processing step. Instead of treating each geo coordinate as its own state, we may group observations together if they fall within some area on a map. Let's say, for example, within one mile of some known landmark. If we create n-number of these boxes to group observations into, we have reduced our state space to be discrete. [7] deals with this problem in a similar manner. We may then apply the same analytical methods detailed in this work to this discretized state space.

### 6.4. Future ontology work

This paper presented an ontological model for Markov probabilities. But we see this model as generalizable to many kinds of probabilities that are calculated using observations of past processes. Future work will explore how the model developed in this paper can be generalized to model Bayesian probabilities, for example.

Future work will also explore how named graphs should be used to model the future-directed representations that are based on Markov and Bayesian probabilities. Such work will include recommendations about how to relate the probabilities to the future-directed representations, like expectations. It will also explore the modeling of the processes that are expected.

Last, future work will concern how to link probabilities to dispositions in BFO conformant OWL ontologies. There is already some work on this in the context of a first-order logic version of BFO that recognizes non-actual instances [6]. The official version of BFO does not recognize non-actual instances. Our work is in OWL and only recognizes actual instances. Thus, we are in a good position to make the insights of [6] implementable in BFO-OWL ontologies.

## 7. Conclusion

In this paper we showed how knowledge graphs structured according to an ontology can be directly accessed to calculate predictions about future processes. We provided two ways to query a knowledge graph for information that can be used to measure the Markov probabilities for fishingVessel to realize dispositions to travel to locations of interest. We then provided the ontology that can be used to structure the information about probabilities, and integrate it back into the knowledge graph. This methodology can be scaled to many similar or dissimilar objects exhibiting the same patterns of behavior. Importantly, the standardized representation of the knowledge in the graph allows us to align our knowledge of the domain with a machine-understandable representation of data so that we can layer additional advanced analytics on top of the knowledge in the graph in a way that will provide an audit trail for techniques that are otherwise considered “blackbox,” rote learning algorithms. This feature of predictive analysis using ontology-based knowledge graphs is important when decisions must be supported by auditable analytics and data that is stored in a standardized, logical construct.